\documentclass[11pt, a4paper]{article} 

\usepackage{graphicx}
\usepackage{vmargin} 

\usepackage{biblatex} 

\bibliography{citation}

\setmargins{2.5cm}{1.5cm}{16.5cm}{23.42cm}{10pt}{1cm}{0pt}{2cm}
\graphicspath{ {./imagenes/} }

\usepackage{fancyhdr}

\begin{document}
\begin{centering}
{ \bfseries \Large RigoBERTa: A State-of-the-Art Language Model For Spanish}
\vspace{0.5cm}

\begin{tabular}{ccc}
 {\small \textbf{Alejandro Vaca Serrano$^1$}} & {\small  \textbf{Guillem García Subies$^1$}} & {\small \textbf{Helena Montoro Zamorano$^1$}}  \\
 {\small \textbf{Nuria Aldama García$^1$}} & {\small \textbf{Doaa Samy$^1$}} & {\small \textbf{David Betancur Sánchez$^1$}} \\
 {\small \textbf{Antonio Moreno Sandoval$^{1,2}$}} &  {\small \textbf{Marta Guerrero Nieto$^1$}} &  {\small \textbf{Álvaro Barbero Jiménez$^{1,2}$}}
\end{tabular}
\begin{center}
    \normalfont{$^1$Instituto de Ingeniería del Conocimiento} \\
    \normalfont{$^2$Universidad Autónoma de Madrid} \\
    \normalfont{\{name.firstsurname\}@iic.uam.es} \\
\end{center}
\vspace{1cm}
{\bfseries \large Abstract} \\
\vspace{0.5cm}
\end{centering}
\parbox{15cm}{\small This paper presents RigoBERTa, a State-of-the-Art Language Model for Spanish. RigoBERTa is trained over a well-curated corpus formed up from different subcorpora with key features. It follows the DeBERTa architecture, which has several advantages over other architectures of similar size as BERT or RoBERTa.
\par
RigoBERTa performance is assessed over 13 NLU tasks in comparison with other available Spanish language models, namely, MarIA, BERTIN and BETO. RigoBERTa outperformed the three models in 10 out of the 13 tasks, achieving new "State-of-the-Art" results.
}

\section{Introduction}

The field of Natural Language Processing (NLP) has witnessed significant advances in the last years, specially with the use of the Transformer architecture, an encoder-decoder model based on the Attention Mechanism \cite{vaswani}. Several models, such as BERT \cite{devlin19}, an encoder-based model useful for language understanding tasks, or GPT-2 \cite{radford}, a decoder-based model specialized on language generation tasks, are based on this architecture. More recently, extremely large language models such as GPT-3 \cite{gpt3} have pushed the State-of-the-Art of NLP to previously unknown limits.

These language models introduce unsupervised pre-training methods that make use of huge amounts of raw text corpora to generate not just word embeddings, but a complete contextualization pipeline for creating general, effective and meaningful representations of language, which is useful for solving different NLP tasks. Due to these recent developments, the leading strategy for good performance is to first pre-train a language model with huge amounts of unlabeled data, and then fine-tune the language model for specific tasks with the use of labeled data.


The most remarkable improvements in this field have been carried out in the English language, for which there are many pre-trained, freely available models. More recently, there has been a growing effort for applying these techniques to other languages, such as French \cite{martin20} \cite{Le20} \cite{lepetit}, Dutch \cite{Delobelle20}, German \cite{gottbert} \cite{germanbert} \cite{moregermanmodels} and many others.

In spite of being one of the most spoken languages worldwide, there are not many language models available for the Spanish language. However, in the last year, more models have been released. Unfortunately, they are not as effective as English language models, due to the following reasons:

\begin{enumerate}
\item Scarcity of Spanish corpora of the same quality and volume as those used by English models.
\item Costs for pre-training large language models are substantial, in the order of hundreds of thousands of dollars, affordable to large multinational companies. For this reason the majority of large language models are in English, as it is the language in which the most economic benefit can be obtained. Most language models in other languages have been developed within the academia.
\end{enumerate}
We aim to overcome those challenges by developing RigoBERTa, a State-of-the-Art language model in Spanish. To that end, a big quality corpus was developed and the selected model architecture achieves State-of-the-Art results faster than preceding architectures such as BERT \cite{devlin19} or RoBERTa \cite{roberta}. Other improvements to previous methods were also implemented, which will be described throughout the article, carried out by a multidisciplinary team. We will describe the techniques used for overcoming the aforementioned difficulties further.

The article is structured in sections. Section \ref{relatedwork} reviews the main advances in the field regarding the Spanish language. The methodology is described in Sections \ref{rigocorpussection} and \ref{rigobertatraining}. In section \ref{experiments}, fine-tuning experiments on 13 NLU tasks are described. Finally, results obtained are reported and analyzed highlighting the progress achieved by RigoBERTa over the three Spanish models (MarIA, BETO, BERTIN).

\section{Related Work}\label{relatedwork}

In this section, recent advances in language models are reviewed. The scope of this review is limited to the Spanish language models. The first publicly available monolingual model in Spanish was BETO \cite{beto}, a Spanish BERT model. The authors implement improvements from RoBERTa \cite{roberta}, such as dynamic masking and whole-word masking, which were also applied to BERT as an update to the original paper \cite{devlin19}. Only base versions of the model are released, with 110M parameters, and a vocabulary size of 32k. Two models were published, cased and uncased.

Both models are able to outperform multilingual BERT \cite{devlin19}\footnote{Multilingual BERT \cite{devlin19} was trained on the top 104 languages with the largest Wikipedia, using the same architecture and training procedure as the original BERT model for English.} on different benchmarks, thus showing that a monolingual model with enough training can outperform a multilingual model, even when more resources and training are used for the latter. The corpus used for training this language model was SUC (Spanish Unnanotated Corpora) \cite{suc}, which is about 18GB of raw texts from different sources such as Spanish Wikipedia or the Spanish portion of OpenSubtitles2018.

Training is carried out in two steps \cite{beto}, the first 900k steps with batch size of 2048 and sequence length of 128. The second consist of 1.1M steps with batch size of 256 and maximum sequence length of 512. The authors also contribute to the establishment of a baseline for evaluating language models in Spanish, by selecting a variety of Natural Language Understanding (NLU) tasks as a benchmark for Spanish models \cite{beto} similar to the English GLUE Benchmark \cite{GLUE}.

More recently, two more language models for Spanish have been developed. Both of them are versions of the RoBERTa \cite{roberta} architecture.

The first one, called BERTIN \cite{bertinpaper}, was developed as part of the Flax/JAX Community Event. Although they faced time and resources limitations, they were able to develop the project without incurring in hardware costs. For BERTIN, as mentioned, a RoBERTa architecture was chosen, with a vocabulary size of 50,265 (same as \cite{roberta}).

A filtered version of MC4 is used, introduced in MT5 paper \cite{mt5}, a multilingual version of T5 \cite{t5}. As the original Spanish portion of MC4 was too big to handle for their computing resources and time, around 50M documents were sampled from the original ~400M documents forming the corpus, with a technique called perplexity-sampling \cite{bertinpaper}.


The model was trained for 250K iterations. For the first 230K steps a sequence length of 128 was used and the last 20K steps sequence length was increased to 512, similarly to BETO \cite{beto}.

MarIA \cite{bsc}, the latest language model released in Spanish, was developed in the context of the \emph{Plan de Impulso de Tecnologías del Lenguaje}\footnote{\url{https://plantl.mineco.gob.es/Paginas/index.aspx}}, funded by the Spanish Government, in collaboration with the Barcelona SuperComputing Center\footnote{\url{https://www.bsc.es/}} and \emph{Biblioteca Nacional de España}\footnote{\url{http://www.bne.es/es/Inicio/index.html}}.

The corpus was 570GB of cleaned texts (see \cite{bsc} for more details on their cleaning process), bigger than any of the corpora used by the rest of the language models in Spanish. A RoBERTa model \cite{bsc} was developed, with a vocabulary size of 50,262. The number of training steps is not specified, albeit the model is trained for one complete epoch. The training process lasted 48 hours with 16 computing nodes, each one with 4 NVIDIA V100 GPUs of 16GB VRAM.


Hyperparameters such as batch size or learning rate have not been reported, and the training corpus has not been published. Although a large version of the model was also developed, it is not considered in this work, as similar results are reported compared to the base one, and the objective of this work is to compare RigoBERTa against models of similar size.

Another contribution of \cite{bsc} was to further extend the benchmark created first by \cite{beto}, by adding more language understanding tasks. Crucially, SQAC (Spanish Question Answering Corpus)\footnote{\url{https://huggingface.co/datasets/PlanTL-GOB-ES/SQAC}} was introduced, the first Question Answering corpus developed originally in Spanish\footnote{the other existing datasets of this type are mainly built by automatic translations}. This is very useful not only because we can have a more complete overview on the models' performance, but also because it enables developing high quality Question Answering models in Spanish.

BETO \cite{beto}, BERTIN \cite{bertinpaper}, and the models presented in the paper, called MarIA \cite{bsc}, were evaluated against the presented benchmark. An effective hyperparameters space was published, easing the future task of evaluating these models against new Spanish language models. The reported results show that MarIA is the best performing model on a majority of tasks, while BETO \cite{beto} performs equal or better on some tasks, and BERTIN \cite{bertinpaper} would be in third position.



\section{Pre-training data}\label{rigocorpussection}

As mentioned above, one of the key aspects for training a proper language model is the corpus quality, as it contains all the information or knowledge the model will be able to store and learn. An ideal corpus for training a general-domain language model should be large, representative, varied and clean, as in \cite{roberta} \cite{devlin19} \cite{mt5} \cite{t5} \cite{deberta} \cite{radford}.

With these basic principles in mind, several different Spanish corpora were used, using a variety of cleaning techniques. Both of these steps are carefully explained in the following lines. 

\subsection{OSCAR: Open Super-large Crawled ALMAnaCH coRpus}\label{oscarsubs}

OSCAR \cite{oscar1} \cite{oscar2} is a very large multilingual corpus, obtained by language classification and filtering of the CommonCrawl\footnote{\url{https://commoncrawl.org/}}. It has a portion of Spanish, of about 149GB of fuzzily and superficially deduplicated texts\footnote{Obtained from \url{https://huggingface.co/datasets/oscar}}.

As with the rest of the corpora, the linguists' team analyzed some random portions of Spanish OSCAR. This preliminary analysis releaved that some parts of the corpus were corrupt and do not represent the actual use of Spanish language. For example, javascript code, automatic writings, poor automatic translations or malformed sentences, as expected from CommonCrawl corpora.

The linguistics team developed a set of quality guidelines to determine whether a text resembles any possible human use of the language, to only include texts for training that meet these criteria. Following these guidelines, a 5,000 texts dataset with valid/non-valid labels was created for later use.


\subsection{Internal Sources}
We had access to a news source through an official provider. This provided us with 70 GB of raw texts of varying quality, according to the linguistics team. Although most news articles seemed of good quality, other articles looked more likely to be corrupt or exhibited a bad use of language (repetitive meaningless texts, texts full of punctuation marks, etc.). However, being these a minority, the corpus passed the quality check for being part of the training data for RigoBERTa.

\subsection{SUC: Spanish Unnanotated Corpora}

SUC \cite{suc} is a corpus comprised of many different small subcorpora, taken from the Huggingface Datasets Library \cite{hfdatasets} \footnote{\url{https://huggingface.co/datasets/large_spanish_corpus}}. It is a relatively small corpus, as compared to the rest of the corpora used for pre-training RigoBERTa, with only 18GB of raw texts. Moreover, most of its texts are too short and documents are splitted in paragraphs and shuffled (this makes it impossible to rebuild the sentences into the original documents, thus the model could not see complete contexts.). Because of this, its already small size was significantly diminished after the cleaning process.

\subsection{mC4}
MC4 \cite{mt5} was included as part of the training data, applying the same approach for filtering as \cite{bertinpaper}, using "preplexity-sampling" technique. This resulted in ~1TB of raw texts, in full documents, around 60\% of the original Spanish portion of MC4. The analysis performed by our team resulted in a similar conclusion to that of OSCAR, that is, around 50\% or more of the texts are of good quality, but a rigorous cleaning must be done in order to obtain a clean version of the corpus. Moreover, the mC4 is formed from crawls from the Internet, as OSCAR and a portion of SUC, therefore we could expect to have many duplicate texts in our dataset, which is another feature to avoid in quality checks.

\subsection{Cleaning Process}

For each of these corpora, we performed the same cleaning process, which consisted on language, length and punctuation filters.
First of all, we used a language detection model for filtering out those texts with very low probability of being written in Spanish. With this we emulate the filtering carried out in \cite{t5} and \cite{mt5}. Among model languages in Spanish, in \cite{bsc} a similar cleaning process was carried out.

Second, documents with less than 200 characters were filtered out following \cite{t5} \cite{mt5}. By applying this filter, most of SUC texts remain excluded.

Third, to fix encoding issues and similar problems ftfy\footnote{\url{https://pypi.org/project/ftfy/}} was used, so that texts originally in latin-1, for example, which had been corrupted when transformed into utf-8, were fixed to their original form.

Since MC4 was still too big, a punctuation filter was carried out. For this, linguistic rules were developed, aimed at removing low quality texts not filtered by the previous processes.


The cleaning process of GPT-3 \cite{gpt3} was replicated. It starts with a deduplicating process with the use of PySpark MinHashLSH\footnote{\url{https://spark.apache.org/docs/2.2.3/ml-features.html\#minhash-for-jaccard-distance}}. Secondly, a set of high quality texts was used to train a logistic regression model to distinguish between good and bad quality texts, then a filtering rule based on the Pareto distribution is applied, as in GPT-3 \cite{gpt3}.

Two experiments were conducted. First, the samples manually annotated as good/bad quality by our team described in \ref{oscarsubs} were used. Secondly, we used the Wikipedia and the good quality part of Internal Sources as positive examples and the rest of the corpus as negative examples.

However, after analyzing the outcomes of each of these experiments, it was decided to keep only the first part of the cleaning process of GPT-3 \cite{gpt3}, that is, the fuzzy deduplication of documents.
After applying all the cleaning processes, we end up with a total volume of Spanish texts for training of similar size, although slightly smaller, to the corpus used by MarIA \cite{bsc}. These training data are of high quality given the strict cleaning filters applied.



\section{RigoBERTa Model and Training}\label{rigobertatraining}

In this section we describe the model architecture, as well as the training process used for obtaining a State-of-the-Art language model in Spanish. We start by explaining the procedure followed for training the tokenizer. Before that, it is important to mention that we use DeBERTa \cite{deberta}, which improves over Bert \cite{devlin19} and Roberta \cite{roberta}, which has an impact on the following subsections.

\subsection{RigoTokenizer}

For the training process ByteLevelBPETokenizer from Huggingface Tokenizers library\footnote{\url{https://huggingface.co/docs/tokenizers/python/latest/}} was used. It is the tokenization algorithm used by DeBERTa \cite{deberta}, the model architecture of RigoBERTa, and very similar to the tokenizer used by MarIA \cite{bsc} and BERTIN \cite{bertinpaper}, with a vocabulary size of 50,265, as stated in \cite{deberta}.

\subsection{Model Architecture}

We chose DeBERTa \cite{deberta} as the model architecture, as the English version clearly outperforms English BERT and RoBERTa \emph{base} models. The authors of the model also show that it converges faster, so it needs less training time and resources than the English RoBERTa model. Therefore, due to its similar size, increased training efficiency and specially, its enhanced effectiveness when fine-tuning, we concluded that this was the best possible model of \emph{base} size for developing RigoBERTa. For more details about the improvements of this model with respect to BERT or RoBERTa, see the original article \cite{deberta}.

\subsection{Training Procedure}



For hyperparameters and training configuration, experiments were carried out on a small subset of the corpus. The final training was then launched with the best configuration from those experiments. The training was carried out by a limited budget of 250k steps, about the same as BERTIN \cite{bertinpaper}, less than one third of the steps used in BETO \footnote{We perform an estimation of the number of steps of BETO with 512 sequence length; as they perform an initial training phase using 128 sequence length, we count each of these steps as 1/4 of a step.} \cite{beto} and between a half and a third of the steps trained by MarIA \cite{bsc}. This is specially relevant when we look at the experimental results on different benchmark tasks. With less training, RigoBERTa was able to achive better results in many NLU tasks, as will be later shown and discussed in the paper.

Regarding the hardware used for training the model, a single AWS p4d.24xlarge instance, with 8 NVIDIA A100 GPUs with 40GB of memory each, was used.

\section{Experimental Setting: Spanish GLUES}\label{experiments}

Building on the work by \cite{beto} and \cite{bsc}, a number of NLU tasks were selected from a business-oriented perspective. With this set of tasks, performance of RigboBERTA could be demostrated statistically and experimentally. The tasks that form the benchmark for evaluating those models are described next.

\textbf{Natural Language Inference: XNLI} \cite{xnli} Natural language inference is the task of determining whether a "hypothesis" is true (entailment), false (contradiction), or undetermined (neutral) given a "premise". MNLI \cite{mnli} is a dataset for such task in English. It was automatically translated to 15 languages, forming the XNLI \cite{xnli}. We use the Spanish portion of this dataset, from Huggingface Datasets library \cite{hfdatasets} \footnote{\url{https://huggingface.co/datasets/xnli}}.

\textbf{Paraphrasing: PAWS-X} \cite{paws-x} is a multilingual version of PAWS \cite{paws} with pairs of sentences which have to be predicted as semantically equivalent or not. We used the Spanish version of the dataset\footnote{\url{https://huggingface.co/datasets/paws-x}}.

\textbf{Document Classification: MLDoc} \cite{mldoc} is a subset of the Reuters corpus\footnote{\url{https://trec.nist.gov/data/reuters/reuters.html}}, which we extracted using Facebook's library MLDoc\footnote{\url{https://github.com/facebookresearch/MLDoc}}. We decided to use both the European Spanish and Latin American Spanish subsets, for training, development and test, getting in each case the biggest version of the dataset available (versions of different sizes are available). The task for this dataset consists on classifying each document inside the following categories: CCAT (Corporate/Industrial), ECAT (Economics), GCAT (Government/Social) and MCAT (Markets).

\textbf{Workshop on Sentiment Analysis at SEPLN: TASS} \cite{tass2020} is a small sentiment analysis dataset of different varieties of Spanish which was part of the SEPLN 2020\footnote{\url{http://sepln2020.sepln.org/}}. We used a compilation of all versions of Spanish for this task.

\textbf{Spanish Question Answering Corpus: SQAC} \cite{bsc} was introduced in section \ref{relatedwork} as part of the contributions by the authors of MarIA. It has a similar format to SQUAD v1.1 \cite{squadv1}, with no adversarial questions, and it is formed with 18,817 questions and 6,247 textual contexts \cite{bsc}. It is freely accessible\footnote{\url{https://huggingface.co/datasets/PlanTL-GOB-ES/SQAC}}.

\textbf{Cross-Lingual Question Answering Dataset: XQUAD + SQUAD-ES} \cite{xquad} \cite{squades}. XQUAD \cite{xquad} is a collection of 1190 question-answer pairs. These pairs were translated by professional human translators to ten languages from the evaluation set of SQUAD v1.1. As it is a very small dataset of well-known quality, we used the Spanish portion of it as the test set\footnote{\url{https://huggingface.co/datasets/xquad}}. A machine-translated version of SQUAD v1.1, SQUAD-ES\footnote{\url{https://huggingface.co/datasets/squad_es}} \cite{squades}, was used for training and development sets. As there is only one split in Huggingface Datasets \cite{hfdatasets} version, we split 10\% of it as the development set.

\textbf{CAPITEL-NER}\footnote{\url{https://sites.google.com/view/capitel2020}} \cite{capitelner} is a NER task which was part of IberLEF 2020\footnote{\url{https://sites.google.com/view/iberlef2020/home}}, focusing on identifying each token as Person (PER), Location (LOC), Organization (ORG) or Other (OTH).

\textbf{ConLL2002} \cite{conll2002} is a NER task including the tags PER, LOC, ORG and MISC\footnote{\url{https://github.com/teropa/nlp/tree/master/resources/corpora/conll2002}}. Although it is an old dataset, it is still widely used for benchmarking models in the NER task in Spanish \cite{beto} \cite{bsc}.

\textbf{CANcer TExt Mining Shared Task – tumor named entity recognition: CANTEMISTNER} \cite{cantemistner} is a track from IberLEF 2020 formed by 3 tasks, of which we just use the NER task, focused on detecting tumor morphology mentions in medical texts on a collection of 3000 clinical cases.\footnote{It can be accessed on the competition page: \url{https://temu.bsc.es/cantemist/}}.

\textbf{Medical Document Anonymization Track: MEDDOCAN} \cite{meddocan} is a track of IberLEF2019 that aimed at anonimizing clinical records. Only the NER task is used, consisting on identifying several labels in clinical records\footnote{See \cite{meddocan} for more information}. However, many labels had vey low frequency, so they were discarded. Concretely, the 15 labels with lowest frequecies were removed. As with the rest of the corpora released by BSC for IberLEF Conferences, it is an open-access dataset\footnote{\url{https://temu.bsc.es/meddocan/}}.

\textbf{MEDical DOcuments PROFessions recognition shared task: MEDDOPROF} \cite{meddoprof} is a track focusing on clinical case reports, part of IberLEF2021. Only NER tasks 1 and 2 were used, so they will be referenced as MEDDOPROF1 and MEDDOPROF2, respectively. MEDDOPROF1 consists on identifying professions (PROFESION) and employment status (SITUACION\_LABORAL), while MEDDOPROF2 has labels: patients (PACIENTE), family member (FAMILIAR), health professional (SANITARIO) or someone else related to the patient (OTROS). It is published in the BSC page\footnote{\url{https://temu.bsc.es/meddoprof/}}. The dataset is originally splitted in train and test, so for the development set we choose 10\% of the train set.

\textbf{Pharmacological Substances, Compounds and proteins and Named Entity Recognition track: PHARMACONER}\footnote{\url{https://temu.bsc.es/pharmaconer/}} \cite{pharmaconer} was part of the BioNLP Open Shared Tasks 2019, and the NER task (track 1, the one we use) consists on identifying the entities NORMALIZABLES, NO\_NORMALIZABLES, PROTEINAS (proteins) and UNCLEAR on medical texts.

\section{Evaluation}\label{evalsection}
The four Spanish language models (RigoBERTa, BETO, BERTIN and MarIA) were evaluated on the tasks introduced in the previous section. For that, the original models authors' recommendations were followed. For MarIA, BERTIN and BETO, we used the hyperparameter space in \cite{bsc}, as it is the most extensive research that has been carried out with these models, and these are the hyperparameters they used for training their own model.

It does make sense that MarIA, BERTIN and BETO use the same hyperparameter setup, as RoBERTa and BERT architectures are fundamentally the same, with the only exception of the embeddings layer, as RoBERTa architecture has a vocabulary size of 50,265 while BERT has 32,000.

On the other hand, we used a different architecture, DeBERTa, which has several variations in the architecture with respect to RoBERTa and BERT \cite{deberta}. These are: disentangled attention, relative positional embeddings on each layer, contrasting with the absolute positional embeddings just in the first layer of BERT and RoBERTa, and the  Enhanced Mask Decoder. The latter accounts for absolute word positions before applying the softmax layer for masked token prediction.

For this reason, for RigoBERTa the hyperparameter space proposed in its original paper \cite{deberta} was used, following the principle of using the author's recommended settings for best optimization of each model for the proposed tasks.

There are some notes on the experiments that should be taken into account. First of all, regarding the Question Answering tasks, much higher results for MarIA were obtained compared to the original authors \cite{bsc}. This was due to a bug we encountered on Huggingface Tokenizers library. This caused that, when relying on Huggingface Transformers library's \cite{hftrans} published and standardized methodology for processing Question Answering datasets, some functionality of BPE tokenizers does not work as intended, causing some examples to be incorrectly processed. This affects MarIA, BERTIN and RigoBERTa, but not BETO, which uses WordPiece\cite{wordpiece}. This, in our opinion, was distorting the reported results of MarIA \cite{bsc} and BERTIN \cite{bertinpaper} on this dataset \cite{bsc}.

In order to solve this bug we developed a novel model-agnostic approach for processing Question Answering datasets, explained in detail in \ref{app:QAmethod}, which does not depend much on tokenizers' functionality\footnote{There are essential requirements, such as including the CLS token at the beggining of texts when inputing them to the model.}, and applied several tests to ensure the answers text spans inputed to the model are exactly the same as the original answers in the dataset.

However, even with this methodology, we could not process Question Answering datasets with BERTIN's tokenizer, as it does not add a CLS token\footnote{CLS token is a special token included as the first token in each sentence\cite{devlin19}} at the beginning of the input ids when tokenizing texts, which is an unavoidable condition for processing Question Answering datasets\footnote{Moreover, the CLS token is of crucial importance, as it is the token we pass to the task layer. When absent, the first token in the sentence is used as the CLS token for Classification tasks. We believe BERTIN's modified tokenizer can be, at least partially, the reason for its underperforming results compared to other models in these type of tasks.}. For that reason, the results for BERTIN on datasets SQAC and SQUAD-ES+XQUAD-ES are set to values slightly lower than the minimum score of the rest of the models in each case. By doing this, it is possible to later perform statistical tests on performance for all models including BERTIN.


Adam optimizer was used for all models \cite{adam}, with parameters $\beta_{1}=0.9, \beta_{2}=0.999, \epsilon=1e-6$, following recommendations in \cite{beto} \cite{devlin19} \cite{roberta} \cite{deberta}. The rest of the optimizer's parameters as the weight decay or the warm-up steps are optimized based on each model's parameter search space. We used a maximum sequence length of 512 for all models on all tasks.

\begin{table}
\centering
\begin{tabular}{ c c c c c c } 
\hline
Dataset &    Metric &  MarIA &  BERTIN &   BETO &  RigoBERTa \\
\hline
CANTEMISTNER        &  F1-Macro &  0.923 &   0.795 &  0.899 &      \textbf{0.933*} \\
CAPITEL             &  F1-Macro &  \textbf{0.878*} &   0.865 &  0.870 &      0.874 \\
ConLL2002           &  F1-Macro &  0.899 &   \textbf{0.901*} &  0.896 &      0.895 \\
MEDDOCAN            &  F1-Macro &  0.841 &   0.722 &  0.847 &      \textbf{0.850*} \\
MEDDOPROF1          &  F1-Macro &  0.807 &   0.710 &  0.805 &      \textbf{0.831*} \\
MEDDOPROF2          &  F1-Macro &  0.785 &   0.442 &  0.818 &      \textbf{0.864*} \\
MLDoc               &  F1-Macro &  \textbf{0.956*} &   0.944 &  0.954 &      \textbf{0.956*} \\
PAWS-x              &  F1-Macro &  0.889 &   0.901 &  0.897 &      \textbf{0.910*} \\
PHARMACONER         &  F1-Macro &  0.571 &   0.471 &  0.614 &      \textbf{0.700*} \\
SQAC                &  F1-Macro &  0.866 &   ---   &  0.762 &      \textbf{0.897*} \\
SQUAD-ES + XQUAD-ES &  F1-Macro &  0.818 &   ---   &  0.756 &      \textbf{0.854*} \\
TASS2020            &  F1-Macro &  \textbf{0.473*} &   0.461 &  0.461 &      0.467 \\
XNLI                &  F1-Macro &  0.816 &   0.794 &  0.817 &      \textbf{0.834*} \\
\hline
\end{tabular}
\caption{Evaluation table of models on test set.}
\label{tab:experimentsresults}
\end{table}
For each model and dataset, we ran experiments with Optuna \cite{optuna}, for the minimum between the total number of possible combinations in the hyperparameter space and 20. Each of these runs is evaluated on the development set. Then, the best performing model is evaluated on the test set, from which the final metrics for that model in that dataset are taken. We use F1-Macro \cite{f1macro} for all tasks. In the case of NER tasks, we compute it by comparing the real and predicted label of each token, with no grouping strategy for subtokens forming a word, as this would have introduced uncertainty on whether the performance differences come from models' performances or from the convenience of chosen aggregation strategies for certain models. These final metrics are reported on table \ref{tab:experimentsresults}, with an asterisk indicating the best performing model for each dataset.

\begin{figure}
    \centering
    \includegraphics[width=0.9\textwidth, scale = 0.8]{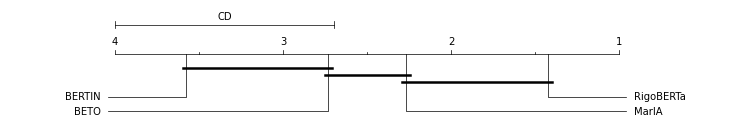}
    \caption{Nemenyi Test for the Language Models in Spanish. This figure represents each model's average rank and the horizontal thick lines group together models whose average ranks are closer to each other than the critical distance, meaning that there is no statistical difference between those models' performances.}
    \label{fig:nemenyi}
\end{figure}

As part of the evaluation, we conducted the Nemenyi Test \cite{nemenyi} \cite{statisticalmodelcomparison} to determine whether there is a statistical difference between models in terms of performance. To do so, the test uses the average rank of each model and the critical distance\footnote{\url{https://edisciplinas.usp.br/pluginfile.php/4129451/mod\_resource/content/1/model\_selection\_evaluation.pdf}}. We have observed a lack of statistical tests on performance for all the rest of the Spanish language models, so we include it to get deeper insights into the effectiveness of each model.

The results from Nemenyi test presented in figure \ref{fig:nemenyi}  show that there is a statistically significant difference between RigoBERTa, with an average rank of 1.42, and both BERTIN, ranking at 3.58, and BETO, with 2.73. Although our model is not proven to be statistically better than MarIA, with an average rank of 2.27, MarIA is not statistically better than BETO, while the latter does not significantly outperform BERTIN.

\subsection{Domain Adaptation: Benchmarking for the Biomedical Domain.}

As an additional experiment, it was decided to compare RigoBERTa against domain-specific models, to assess its domain-transfer ability. The biomedical domain was selected for this, due to the availability of domain-specific models and datasets of reported good quality.

In \cite{bscbio} two domain-specific models are presented. One is for the biomedical domain, while the other is for the biomedical-clinical domain (which is arguably a sub-domain of the former). These two models will be referred to as MarIA-BioMedical and MarIA-BioClinical respectively, as they were developed as part of the MarIA project \cite{bsc} introduced previously. Domain specific corpora were used for pre-training each model, described in detail in \cite{bscbio}. Both use RoBERTa architecture \cite{roberta} \cite{bscbio}.

In \cite{bscbio} both models are evaluated against BETO \cite{beto} on a small set of 3 biomedical and clinical datasets, showing better results than the latter. In this work, MarIA-BioMedical and MarIA-BioClinical are evaluated against MarIA \cite{bsc} and RigoBERTa, being these the two best-performing models in general, as shown in table \ref{tab:experimentsresults}.

Table \ref{tab:experimentsbio} shows the results of each model in terms of f1-score on a set of tasks for the biomedical domain. RigoBERTa achieves the highest f1-score on all tasks, outperforming MarIA, MarIA-BioMedical and MarIA-BioClinical, thus showing a great domain adaptation ability. This is specially relevant because RigoBERTa has not been trained on biomedical texts specifically, contrasting with MarIA-BioMedical and MarIA-BioClinical.

\begin{table}
\centering
\begin{tabular}{ c c c c c c } 
\hline
Dataset &    Metric & MarIA & MarIA-BioMedical &  MarIA-BioClinical &  RigoBERTa \\
\hline
CANTEMISTNER        &  F1-Macro &  0.923 &   0.924 &  0.928 &      \textbf{0.933*} \\
MEDDOCAN            &  F1-Macro &  0.841 &   0.796 &  0.642 &      \textbf{0.850*} \\
MEDDOPROF1          &  F1-Macro &  0.807 &   0.797 &  0.746 &      \textbf{0.831*} \\
MEDDOPROF2          &  F1-Macro &  0.785 &   0.754 &  0.764 &      \textbf{0.864*} \\
PHARMACONER         &  F1-Macro &  0.571 &   0.570 & 0.322 &      \textbf{0.700*} \\
\hline
\end{tabular}
\caption{Evaluation table of models on test set of Biomedical Tasks.}
\label{tab:experimentsbio}
\end{table}

\section{Conclusions \& Future Work}

In this work we presented RigoBERTa, a State-of-the-Art model outperforming the rest of the Spanish language models, as shown in section \ref{evalsection}. In fact, to the best of our knowledge this is the first work reporting statistical tests on performance for language models in Spanish, showing that RigoBERTa is statistically better than BETO \cite{beto} and BERTIN \cite{bertinpaper}. Moreover, RigoBERTa outperforms MarIA \cite{bsc} in 9 out of 13 benchmarks (and obtain similar results in another one).

These results confirm that the main objective of this work was reached: to build the best performing \emph{base} model up to date in Spanish by improving the pre-training corpus and by choosing a better model architecture. These two main contributions together can account for the observed difference in performance. This is most remarkable in Question Answering tasks, which was expected since the DeBERTa architecture \cite{deberta} makes better use of spatial information in texts, very useful for this concrete task.

As for future work, RigoBERTa was only trained for 250k steps, which is a fourth of the total training steps planned. Looking at the figures of fine-tuning performance per training steps in \cite{deberta}, it seems that there is room for much improvement for our model, therefore we will continue training it. Moreover, we plan to keep on training language models as the State-of-the-Art evolves and new improved architectures are published, making use of everything learned during the development of RigoBERTa. Additionally, taking advantage of the great ability of RigoBERTa to learn specific domain tasks, domain-specific models will be developed, by adapting the general-domain model.

\newpage

We would like to acknowledge the support of this project by Instituto de Ingeniería del Conocimiento, as without their faith on us and their investment, this project would not have been possible. Also, we would like to mention the Chair of Computational Linguistics (Cátedra de Lingüística Computacional)\footnote{http://catedras.iic.uam.es/catedra-linguistica-computacional/}, for letting us use their machines for the benchmarking experiments.

\appendix
\section*{Appendix A. Question-Answering Processing Method}
\label{app:QAmethod}

Tokenizers from Huggingface had some issues related to BPE Tokenizers, which cause that the correspondence between generated tokens and original characters (offset mapping) in the text do not match exactly. Therefore when this functionality is used to determine the start and end tokens from start and end characters in the original text, the resulting span of text (between start and end token) is not the actual answer. For that reason, when relying on the official methodology for processing question-answering datasets \footnote{https://github.com/huggingface/notebooks/blob/master/examples/question\_answering.ipynb}, some answers are incorrectly processed, thus having some corrupt examples for training and evaluating, therefore the results obtained by models using BPE tokenizers are far from optimal. This affects, in the context of language models in Spanish, RigoBERTa (presented in this paper), MarIA \cite{bsc} and BERTIN \cite{bertinpaper}.

For solving this, a model-agnostic method was developed, which ensures that the models' inputs between the selected start and end tokens correspond to the exact text answer in the original dataset. By doing this, models do not learn on corrupted examples, therefore their performance on Question Answering tasks is optimized. Then, tests are applied to check the aforementioned condition.

\printbibliography

\end{document}